\documentclass[lettersize,journal]{IEEEtran}
\usepackage{amsmath,amsfonts}
\usepackage{algorithmic}
\usepackage{algorithm}
\usepackage{array}
\usepackage[caption=false,font=normalsize,labelfont=sf,textfont=sf]{subfig}
\usepackage{textcomp}
\usepackage{stfloats}
\usepackage{url}
\usepackage{verbatim}
\usepackage{graphicx}
\usepackage{cite}
\DeclareUnicodeCharacter{2212}{-}
\usepackage{xcolor}
\usepackage{booktabs}
\usepackage{multirow}
\usepackage{amssymb}
\usepackage{pifont}

\def\eg{\textit{e.g.}}
\def\ie{\textit{i.e.}}

\def\etal{\textit{et al.}}

\usepackage{newfloat}
\usepackage{listings}

\begin{document}

\title{ Multi-Modality Driven LoRA for Adverse Condition Depth Estimation}

\author{
Guanglei Yang$^1$ \quad Rui Tian$^1$ \quad Yongqiang Zhang$^2$ \quad Zhun Zhong$^3$\quad Yongqiang Li$^1$\quad Wangmeng Zuo$^{1}$\\
$^1$Harbin Institute of Technology, China  \quad $^2$Inner Mongolia University, China \quad $^3$ Hefei University of Technology , China
 \\

}

\markboth{Journal of \LaTeX\ Class Files,~Vol.~14, No.~8, August~2021}%
{Shell \MakeLowercase{\textit{et al.}}: A Sample Article Using IEEEtran.cls for IEEE Journals}

\maketitle
\begin{abstract}


The autonomous driving community is increasingly focused on addressing corner case problems, particularly those related to ensuring driving safety under adverse conditions (e.g., nighttime, fog, rain). 
To this end, the task of Adverse Condition Depth Estimation (ACDE) has gained significant attention.
Previous approaches in ACDE have primarily relied on generative models, which necessitate additional target images to convert the sunny condition into adverse weather, or learnable parameters for feature augmentation to adapt domain gaps, resulting in increased model complexity and tuning efforts.
Furthermore, unlike CLIP-based methods where textual and visual features have been pre-aligned, depth estimation models lack sufficient alignment between multimodal features, hindering coherent understanding under adverse conditions.
To address these limitations, we propose Multi-Modality Driven LoRA (MMD-LoRA), which leverages low-rank adaptation matrices for efficient fine-tuning from source-domain to target-domain.
It consists of two core components: Prompt Driven Domain Alignment (PDDA) and Visual-Text Consistent Contrastive Learning(VTCCL).
During PDDA, the image encoder with MMD-LoRA generates target-domain visual representations, supervised by alignment loss that the source-target difference between language and image should be equal.
Meanwhile, VTCCL bridges the gap between textual features from CLIP and visual features from diffusion model, pushing apart different weather representations (vision and text) and bringing together similar ones. 
Through extensive experiments, the proposed method achieves state-of-the-art performance on the nuScenes and Oxford RobotCar datasets, underscoring robustness and efficiency in adapting to varied adverse environments.
\end{abstract}

\section{Introduction}
\label{sec:intro}

Autonomous driving systems are designed to operate in various real-world conditions~\cite{kong2024robodepth, adabins, yang2021transformers, EVP, ke2024repurposing, hu2023planning}. 
However, a significant challenge that has drawn increasing attention in recent years is addressing corner case scenarios, where driving safety becomes critical under adverse conditions such as nighttime, fog, rain, and snow~\cite{md4all,tosi2024diffusion, gasperini2021r4dyn, defeat, saunders2023self}. 
These adverse conditions not only limit the vehicle’s ability to perceive the environment but also increase the risk of accidents, making robust depth estimation under such circumstances crucial for safe autonomous driving.

One of the primary hurdles in solving this problem lies in the scarcity of high-quality real-world images from these adverse conditions. It is difficult to collect a comprehensive dataset that adequately captures the wide variety of weather conditions vehicles may encounter. Furthermore, the high cost of annotating these real-world images makes relying solely on traditional data collection and labeling methods impractical.
As a result, there is growing interest in alternative approaches that method in the sunny condition can generalize to these difficult conditions without requiring large-scale labeled data~\cite{vidit2023clip, hu2021lora, wang2024sqldepth,guizilini2023towards,bhat2023zoedepth}.

\begin {figure}[!t]
\centering
\includegraphics[width=1.0\columnwidth]{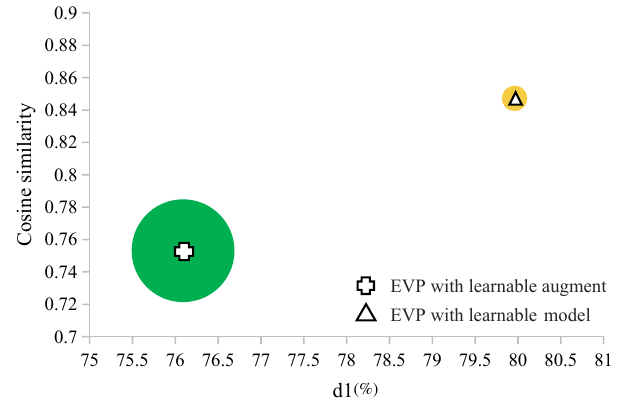}
\vspace{-.3cm}
\caption{Comparison of the baseline depth estimation~\cite{EVP} with learned augmentation and learned LoRA on the nuScenes validation set. 
The y-axis is the cosine similarity between the estimated unseen target-domain visual representation and real target-domain visual representation, and the x-axis denotes the depth estimation performance in $d_1$ for the night scene. 
The bubble size denotes the number of parameters(M). Best be viewed in color and zoomed in.}
\label{tu1}
\vspace{-.3cm}
\end {figure}

To this end, Adverse Condition Depth Estimation (ACDE) has emerged as a promising direction~\cite{yang2024depth,tosi2024diffusion,md4all}. This approach seeks to estimate depth information in unseen weather conditions without relying on a large number of labeled examples. While previous methods, such as md4all~\cite{md4all}, have made notable progress, they predominantly rely on generative models to transform images captured under sunny conditions into those representing adverse weather. However, these generative approaches are dependent on sufficient target images
to build a well-trained model(\eg ForkGAN~\cite{forkgan}).
On the other hand, some methods~\cite{vidit2023clip, saunders2023self, kim2023single} utilize learnable parameters for feature augmentation to adapt to the target domains, resulting in increased model complexity and tuning efforts.
Furthermore, unlike CLIP-based methods~\cite{yu2024turning, CLIP, Wang_2021_CVPR, yarom2024you, xue2023clip} where textual and visual features have been pre-aligned, depth estimation models~\cite{EVP, VPD} lack sufficient alignment to match textual and visual space, hindering coherent understanding under adverse conditions.


In this paper, we present a novel approach to tackle the Adverse Condition Depth Estimation (ACDE) task by introducing Low-Rank Adaptation (LoRA) techniques in conjunction with contrastive learning. 
Specifically, we design Multi-Modality Driven LoRA (MMD-LoRA) to address domain gaps from source-domain to target-domain and misalignment between visual and textual representations.
The core innovation of MMD-LoRA lies in two main components: Prompt Driven Domain Alignment (PDDA) and Visual-Text Consistent Contrastive Learning (VTCCL). 
Prompt Driven Domain Alignment (PDDA) leverages text embeddings to guide the estimation of semantically related visual representations \cite{vidit2023clip, ramesh2022hierarchical}. 
In PDDA, trainable low-rank decomposition matrices are integrated into specific layers of the image encoder and supervised with an alignment loss to maintain consistency between language and image representations across source and target domains. 
This design allows the image encoder with low-rank decomposition matrices to capture accurate target-domain visual features via language knowledge without additional target images.
Compared to the learned augmentation-based method~\cite{vidit2023clip}, our PPDA as a learned model-based method achieves superior depth estimation performance using only a minimal increase in parameter count (0.035M), as illustrated in Figure~\ref{tu1}.
Visual-Text Consistent Contrastive Learning (VTCCL) is designed to align the pre-trained text encoder (based on CLIP) and the pre-trained image encoder (based on the diffusion model). 
By calculating similarity scores between visual representations and text embeddings for each weather condition, VTCCL separates representations of distinct weather conditions while drawing similar ones closer together. This alignment reinforces coherent, consistent representations across modalities, enhancing the generalization ability of MMD-LoRA to varied adverse conditions.

Our contributions are summarized as follows:

\begin{itemize}

\item We present MMD-LoRA, a novel approach to ACDE that efficiently addresses domain gaps and multimodal misalignment by leveraging Low-Rank Adaptation (LoRA) techniques in conjunction with contrastive learning.

\item We propose Prompt Driven Domain Alignment (PDDA), which uses trainable low-rank adaptation matrices within the image encoder guided by text embeddings. This component captures accurate target-domain visual features without the need for additional target images. 
Meanwhile, Visual-Text Consistent Contrastive Learning (VTCCL) is designed to achieve robust multimodal alignment, reinforcing consistent representations by separating embeddings of different weather conditions while drawing similar ones together.

\item Extensive experiments demonstrate the effectiveness of MMD-LoRA for adverse condition depth estimation on two popular benchmarks, including nuScenes dataset and Oxford RobotCar dataset.
\end{itemize}

\section{Related Work}
\label{sec:Related Work}

\subsection{Adverse condition depth estimation} 

Adverse weather conditions can lead to erroneous measurements from LiDAR sensors, particularly due to reflections on flooded roads during rainy days and non-textured areas illuminated at night. These factors hinder the establishment of accurate depth estimation in pixel correspondences. To date, only a limited number of studies have explored depth estimation in adverse weather conditions~\cite{md4all, liu2021self, tosi2024diffusion, gasperini2021r4dyn, costanzino2023iccv, Wang_2021_ICCV, Saunders_2023_ICCV}.

Recent advances in adverse condition depth estimation have been
made by image augmentation-based methods ~\cite{liu2021self, gasperini2021r4dyn,costanzino2023iccv, Wang_2021_ICCV, Saunders_2023_ICCV} and style transfer-based methods~\cite{md4all, tosi2024diffusion}. 
These methods based on image augmentation tackle depth estimation in adverse weather focusing solely on issues related to poor illumination and reflections. 
However, these methods often fail to establish a unified framework that can deliver a more robust and general solution.
To fill this limitation,  style transfer-based methods are proposed to construct a unified framework for varied adverse weather. 
For example, md4all~\cite{md4all} transforms images captured under sunny conditions into those depicting adverse weather by utilizing generative models like ForkGAN~\cite{forkgan} to diversify source-domain images.
Similarly, Fabio~\etal~\cite{tosi2024diffusion} leverage cutting-edge text-to-image diffusion models to generate new, user-defined scenes along with their associated depth information.

In contrast to previous methods that require extra target images, which limit a vehicle's ability to perceive unseen environments, our proposed method effectively addresses this challenge by 
Low-Rank Adaptation (LoRA) techniques in conjunction with contrastive learning. 
The proposed method enables us to construct a unified framework for each weather condition without the necessity of source-target image pairs. Importantly, our method operates in the feature space rather than the image space. When faced with rare weather conditions(\eg freezing rain, sandstorms), we can simply introduce the corresponding target-domain text descriptions to further optimize MMD-LoRA, thereby eliminating the need to collect source-target image pairs.

\subsection{Zero-Shot Depth Estimation} 

Zero-shot depth estimation~\cite{ranftl2020towards, guizilini2023towards, piccinelli2024unidepth, yin2023metric3d, zhang2022can, yang2024depth, bhat2023zoedepth}
presents a significant challenge task, requiring a depth estimator trained on source-domain images to generalize effectively to an unseen target domain during inference.
For instance, Zoedepth~\cite{bhat2023zoedepth} combines relative and metric depth by pre-training on various datasets and employs a lightweight decoder to fine-tune the model with metric depth information, achieving impressive generalization performance.
Ranftl~\etal~\cite{ranftl2020towards} propose a robust training objective that is invariant to changes in depth range and scale to improve generalization performance by combining data from different sources.
More recently, Depth Anything~\cite{yang2024depth} improves the model generalization 
using an expanded training set to approximately 62 million images. 
Even though these efforts have bolstered zero-shot inference capabilities, there remains a significant need for more high-quality synthetic real-world images.

When high-quality auxiliary or synthetic datasets cannot be obtained, 
the generalization ability to perceive the unseen environment of existing models is limited. 
Thus, we desire to explore the adverse condition depth estimation task without the need for additional auxiliary or synthetic datasets. 
Inspired by the generalization ability of LoRA~\cite{hu2021lora} in the NLP field, 
the trainable low-rank decomposition matrices are integrated into specific layers of Transformer to fine-tune large language models (LLMs) for downstream tasks.
In the papers, Prompt Driven Domain Alignment(PDDA) captures the unseen target-domain visual representation based on the guide of text embeddings to semantically related visual representations.
The PDDA not only improves the generalization ability but also introduces a few trainable parameters.



\subsection{Multi-modality alignment strategy} 

Multi-modality  alignment~\cite{yu2024turning, CLIP, Wang_2021_CVPR, yarom2024you, xue2023clip, ali2023clip, wu2023cora, moayeri2023text, zhou2022extract, zhang2023controlvideo} enhances a model’s scene-aware ability and captures fine-grained representations of real-world scenes. 
For example, Alec Radford~\etal~\cite{CLIP} pioneer the use of natural language as a supervision signal for image representation, aligning visual and text encoders. 
Yu~\etal~\cite{yu2024turning} develop an instance-language matching network that employs visual prompt learning and cross-attention in the CLIP backbone to facilitate the matching of instance and text embeddings.
Zhou~\etal~\cite{zhou2022extract} introduce pseudo labeling and self-training process to achieve the pixel-text alignment for semantic segmentation task in the absence of annotations.

Unlike these CLIP-based methods where textual and visual features have been pre-aligned, depth estimation models lack sufficient
alignment between multimodal features, hindering coherent understanding under adverse conditions.
The misalignment between text encoder and image encoder inevitably disrupts the LoRA generalization ability to adverse conditions and leads to sub-optimality.
To address this issue, the proposed Visual-Text Consistent Contrastive Learning (VTCCL) pushes apart different weather representations and brings together similar ones, enhancing the model’s generalization to varied adverse conditions. 
Based on the training paradigm following vanilla CLIP~\cite{CLIP}, we effectively align consistent representations across modalities in the depth estimation without introducing the additional parameters compared with the first component PDDA.


\section{Method}
\label{sec:method}

\begin{figure*}[!t]
\centering
\includegraphics[width=0.95\textwidth]{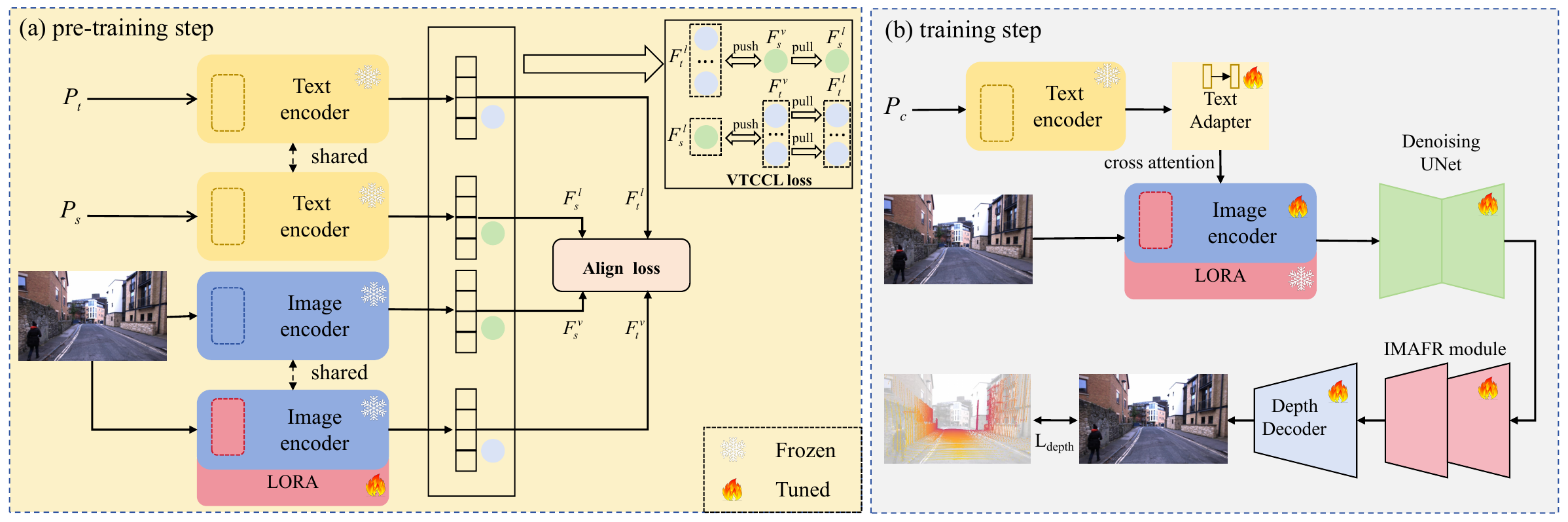}
\vspace{-0.3cm}
\caption{The overview of the MMD-LoRA pipeline including the pre-training step and training step. In the pre-training step, MMD-LoRA captures accurate target-domain visual features and achieves robust multimodal alignment based on the multi-modal learning and contrastive learning paradigm. 
In the training step, we freeze the trained LoRA to inject low-rank decomposition matrices into ’q’,’k’,’v’, ’proj’ layers of the image encoder self-attention and further optimize depth estimator based on the ground-truth depth map.
}
\label{pipeline}
\vspace{-.3cm}
\end{figure*}

As previously discussed, our MMD-LoRA aims to solve adverse condition depth estimation by Prompt Driven Domain Alignment (PDDA) and Visual-Text Consistent Contrastive Learning (VTCCL). 
An overview of our pipeline is depicted in Figure~\ref{pipeline}, including a pre-training step and a training step. 
In the pre-training step, the image encoder in the baseline depth estimator with MMD-LoRA captures accurate target-domain visual representations supervised by alignment loss during PDDA. 
Meanwhile, VTCCL separates representations of distinct weather conditions while drawing similar ones closer together to further enhance the MMD-LoRA generalization to varied adverse conditions.
In the training step, we use the trained MMD-LoRA to inject trainable low-rank decomposition matrices into ’q’,’k’,’v’, ’proj’ layers of self-attention in the image encoder in the depth estimator and further optimize the depth estimator. 

Given the image $I \in \mathbb{R}^{HWC}$ in the source domain, we take randomly multiple crops $I_{crop} \in \mathbb{R}^{N \times H'W'C}$ to perceive the fine-grained weather condition, where $N$ is crops number.
We define the source-domain text description $P_s$ and unseen target-domain text description $P_t= \left\{P_t^i \right\}_{i=1}^{M}$,
where $i$ and $M$ denote the index of target-domain and total number of target-domains. For example, $P_s$ and $P_t^i$ can be denoted as 'an image taken during the day' and 'an image taken on a night', 'an image taken on a rainy day'. 
To diversify the unseen target-domain text description during PDDA, we also combine semantic concepts of multiple adverse weather conditions, such as $P_t^i$ is denoted as 'an image taken on a rainy night'. 
The pre-trained CLIP text encoder $T(\cdot)$ and the pre-trained image encoder $V(\cdot)$ in baseline depth estimator are introduced to extract the visual representation and text embedding. 
Then, our objective is to capture the unseen target-domain visual representation $F_t^{v}$ using parameter-efficient LoRA benefiting from its inherent advantages. 
Following the training paradigm of existing depth estimation~\cite{EVP}, the depth estimator utilizes image caption $P_{c}$ to capture the semantic information of the complex road scenario.

\subsection{Prompt Driven Domain Alignment}
\label{LoRA-based Alignment}

The pre-trained image encoder in existing depth estimators generally captures visual representations under sunny conditions, which limits its ability to generalize to unseen target domains. 
To overcome this limitation, the goal is to capture more robust visual representations and effectively adapt to the baseline depth estimator for adverse condition depth estimation with minimal modifications. 
Inspired by the strong generalization capabilities of Low-Rank Adaptation (LoRA) demonstrated in the NLP domain \cite{hu2021lora}, trainable low-rank decomposition matrices (referred to as MMD-LoRA) are introduced into specific layers of the image encoder.
This integration significantly improves the model's ability to generalize across diverse adverse weather scenarios. Rather than updating the entire vanilla image encoder, modifications are constrained by injecting trainable low-rank matrices, as defined in Eq.~\ref{gongshi0}:

\begin{equation}
\label{gongshi0}
\begin{aligned}
\displaystyle 
W = W_0+\Delta W =  W_0 + BA
\end{aligned}
\end{equation}
where the pre-trained weight matrix of the image encoder is denoted as $W_0 \in \mathbb{R}^{d \times k}$.
The $\Delta W$ is constructed by the trainable low-rank decomposition matrices, B and A, to support transfer ability from the sunny condition to adverse weather, where $B \in \mathbb{R}^{d \times r}$, $A \in \mathbb{R}^{r \times k}$, and the rank $r \ll min(d, k)$. 
The trainable low-rank decomposition matrices B and A are designed to capture the various unseen target-domain visual representations.



To train two matrices(\ie B and A), PDDA leverages text embeddings to guide the estimation of semantically related visual representations on unseen target-domain~\cite{vidit2023clip, ramesh2022hierarchical}.
Specifically, the text encoder $T(\cdot)$ and the image encoder $V(\cdot)$ capture source-domain visual representations $F_s^v$, source-domain text embeddings $F_s^l$ and target-domain text embeddings $F_t^{l}$ during the pre-training step.
The unseen target-domain visual representations $F_t^v$ are calculated by Eq.~\ref{gongshi3}:
\begin{equation}
\label{gongshi3}
\begin{aligned}
\displaystyle 
F_t^v = [V \odot LoRA](I_{crop})
\end{aligned}
\end{equation}
where $\odot$ denotes the trainable low-rank decomposition matrices(\ie MMD-LoRA) are integrated into 'q', 'k', 'v' and 'proj' layer of self-attention in the image encoder $V(\cdot)$.
In the textual and visual space, the source-target difference between language and image should be equal. 
Thus, the text difference $\Delta L$ helps the estimation of the visual difference $\Delta V$ between the source domain and target domain. 
We design an alignment loss to supervise the MMD-LoRA for further obtaining accurate unseen target-domain visual representation by Eq.~\ref{gongshi2}:
\begin{equation}
\label{gongshi2}
\begin{aligned}
\displaystyle 
L_{align}=D(\Delta V, \Delta L)+||F_t^v-F_s^v||_1
\end{aligned}
\end{equation}
where the $D(\cdot)$ denotes the cosine similarity loss. $||\cdot||_1$ denotes the L1 regularizer to prevent the predicted unseen target-domain representation from deviating too far from the source-domain visual representation. 
$\Delta V$ denotes the difference of $F_s^v$ and $F_t^v$ obtained by Eq.~\ref{gongshi3}. 
$\Delta L$ denotes the difference of $F_s^l$ and $F_t^l$ by text encoder $V(\cdot)$. 
Note that the text encoder $T(\cdot)$ and image encoder $V(\cdot)$ are frozen.


\subsection{Visual-Text Consistent Contrastive Learning}

Based on the above process, PDDA enables the image encoder with MMD-LoRA to capture accurate target-domain visual features $F_t^{v}$ for depth estimation without requiring access to target-domain images. 
However, existing depth estimation models lack sufficient alignment between multimodal features which will inevitably disrupt the MMD-LoRA generalization ability, leading to a sub-optimal problem.
To solve the sub-optimal problem, we design Visual-Text Consistent Contrastive Learning (VTCCL) achieving robust multimodal alignment and further improving MMD-LoRA generalization ability for adverse weather.

Specifically, for each image crops $I_{crop}$ in the image $I$, the $F_s^v$, $F_s^l$, $F_t^v$ and $F_t^l$ of each crop are utilized to calculate the similarity of visual representations and text embeddings for each weather.
We iteratively push apart different weather representations (vision and text) and bring together similar ones to further enhance MMD-LoRA generalization ability to adverse conditions by Eq.~\ref{gongshi5} following the training paradigm of contrastive learning.
\begin{equation}\small
\label{gongshi5}
\begin{aligned}
\displaystyle 
L_{vtccl}=-\lambda_0 \cdot log\frac{exp(F_s^{v}\cdot F_s^{l}/\tau)}{exp(F_s^{v}\cdot F_s^{l}/\tau)+\sum_{i=1}^{M} exp(F_s^{v}\cdot F_t^{l_i}/\tau)}-\\[1em]
\sum_{i=1}^{M} \lambda_i \cdot log \frac{exp(F_{t}^{v_i}\cdot F_{t}^{l_i}/\tau)}{exp(F_t^{v_i}\cdot F_t^{l_i}/\tau)+exp(F_t^{v_i}\cdot F_s^{l}/\tau)}
\end{aligned}
\end{equation}
where $F_s^v$, $F_t^{v_i}$  $\in$ $\mathbb{R}^{N \times D}$ and $F_s^{l}$, $F_t^{l_i}$ $\in$ $\mathbb{R}^{D}$. 
The i, N and M denote the index of adverse weathers, the number of image crops and the total number of adverse weathers, respectively. 
The $\lambda$ denotes the contrastive learning weight coefficient controlling different weights of each weather in Eq.~\ref{gongshi5}.
The $\tau$ denotes a temperature parameter. 
For a brief, we omitted the N sum symbol of the visual and textual representation for all image crops.

For a better understanding of Eq.~\ref{gongshi5}, we consider an example with a generic text description $P_s$ for the source domain, such as "an image taken during the day." 
For the target domain, a set of text descriptions, $P_t= \left\{P_t^i \right\}_{i=1}^{M}$, representing adverse conditions like "an image taken on a night" and "an image taken on a rainy day."
In this context, M is set to 2 for simplicity.
These text descriptions are then processed using the pre-trained CLIP text encoder, resulting in source and target text embeddings, $F_s^l$ and $F_t^l$, respectively.
 

Eq.~\ref{gongshi5} applies two contrastive learning terms: one focuses on sunny conditions, while the other addresses adverse weather (e.g., night, rain).
For the sunny condition contrastive learning term, the source-domain visual representation $F_s^v$ is selected as the anchor. The source-domain text embeddings $F_s^l$ are treated as positive samples, while the target-domain text embeddings $F_t^l$ serve as negative samples. The similarity is calculated in the projection space using inner products to measure the distance between visual representations and text embeddings.
In the case of adverse weather contrastive learning, a similar approach is adopted. For instance, the night-domain visual representation $F_t^{v_1}$ is chosen as the anchor, with the corresponding night-domain text embedding $F_t^{l_1}$ assigned as a positive sample and the source-domain text embedding $F_s^l$ designated as a negative sample. This process is repeated iteratively for all target-domain conditions, including night and rainy.

Finally, the overall loss in the pre-trained stage is calculated by Eq.~\ref{gongshi6}.
\begin{equation}
\label{gongshi6}
\begin{aligned}
\displaystyle 
L_{pre} = L_{align}+L_{vtccl}
\end{aligned}
\end{equation}



\subsection{Training architecture}

After the pre-training step, the trained MMD-LoRA is injected into ’q’,’k’,’v’, ’proj’ layers of the self-attention in the image encoder and we further optimize the baseline depth estimator(\eg EVP~\cite{EVP}) by ground-truth depth map.
Specifically, training step as shown in Figure~\ref{pipeline}, the baseline depth estimator utilizes image caption $P_{c}$ to capture the semantic information of the complex road scenario, and the image encoder with trained MMD-LoRA captures the source-domain visual representation and unseen target-domain visual representations.
The various visual representations and text embedding of image captions utilize cross-attention maps to provide explicit guidance for depth estimation tasks. 
Then, the denoising U-Net’s decoder features feed on the Inverse Multi-Attentive Feature Refinement (IMFAR) to capture enhanced visual representation and the depth decoder outputs the predicted depth map. The training step is described in detail in the Supplementary.


\begin{table*}[t]
\scriptsize
\centering
\caption{Evaluation on the nuScenes validation set. 
The bold entries denote that our MMD-LoRA surpasses previous SOTA results.}
\vspace{-0.2cm}
\renewcommand{\arraystretch}{1.3}
\begin{tabular}{l c c c c c c c c c}
\toprule
\multirow{2}{*}{Method} & \multicolumn{3}{c}{day-clear-nuScenes} 
&\multicolumn{3}{c}{night-nuScenes} &\multicolumn{3}{c}{day-rain-nuScenes} \\ \cmidrule(r){2-4}  \cmidrule(r){5-7} \cmidrule(r){8-10}
& absREL($\downarrow$) & RMSE($\downarrow$) & $d_1$($\uparrow$) & absREL($\downarrow$) & RMSE($\downarrow$) & $d_1$($\uparrow$) & absREL($\downarrow$) & RMSE($\downarrow$) & $d_1$($\uparrow$) \\ 
\midrule
Monodepth2~\cite{godard2019digging}	&0.1333	&6.459	&85.88	&0.2419	&10.922 &58.17	&0.1572	&7.453	&79.49 \\ 
PackNet-SfM~\cite{guizilini20203d} 	&0.1567	&7.230 	&82.64	&0.2617	&11.063	&56.64	&0.1645	&8.288	&77.07 \\
R4Dyn w/o r	in~\cite{gasperini2021r4dyn} 	&0.1296	&6.536	&85.76	&0.2731	&12.43	&52.85	&0.1465	&7.533	&80.59 \\ 
R4Dyn(radar)~\cite{gasperini2021r4dyn}		&0.1259	&6.434	&86.97	&0.2194	&10.542	&62.28	&0.1337	&7.131	&83.91\\
RNW~\cite{yang2024self}	&0.2872	&9.185	&56.21	&0.3333	&10.098	&43.72	&0.2952	&9.341	&57.21\\
DPT~\cite{ranftl2020towards} &0.1890 &8.094 &75.39 &0.3540 &12.875 &60.97 &0.2370 &8.780 &66.96\\
AdaBins~\cite{adabins}	&0.1384	&5.582	&81.31	&0.2296&7.344	&63.95	&0.1726	&6.267	&76.01\\
Depth Anything~\cite{yang2024depth}
&0.1340 &6.792 &82.53 &0.2190 &9.140 &70.26 &0.1570 &7.570 &77.42\\
md4all-DD~\cite{md4all}	&0.1366	&6.452	&84.61	&0.1921	&8.507	&71.07	&0.1414	&7.228	&80.98\\
DepthAnything with ft.~\cite{tosi2024diffusion} &0.1340 &6.792 &82.53 &0.2190 &9.140 &70.26 &0.1570 &7.570 &77.42\\
md4all-AD~\cite{md4all}	&0.1206	&4.806	&88.03	&0.1821	&\textbf{6.372}	&75.33	&0.1562	&5.903	&82.82\\
MMD-LoRA(ours)	&\textbf{0.0690} 	&\textbf{3.192}	&\textbf{96.46}	&\textbf{0.1545}	&7.127	&\textbf{79.96}	&\textbf{0.0740} 	&\textbf{3.417}	&\textbf{95.37} \\
\bottomrule
\end{tabular}
\label{main_tab_nus}
\vspace{-0.2cm}
\end{table*}

\section{Experiments}
\label{sec:exp}

\begin{figure*}[!t]
\centering
\includegraphics[width=0.9\textwidth]{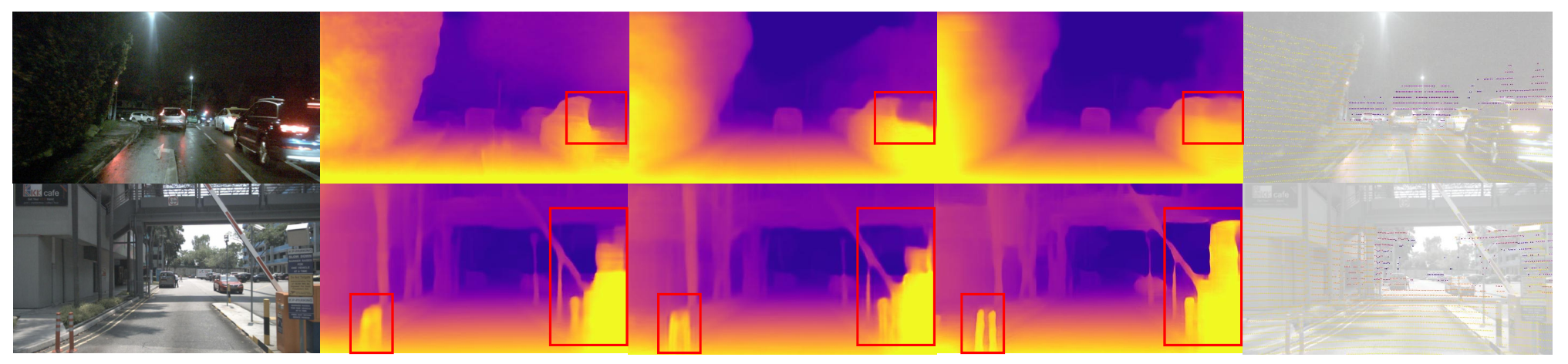}
\vspace{-0.3cm}
\caption{Qualitative results of our proposed MMD-LoRA and the previous SOTA depth estimation method on the nuScenes validation sets. 
The first column denotes the original image. The second, third and fourth column denote the depth estimation results of Monodepth2, md4allDD and ours MMD-LoRA respectively. 
The final column indicates the ground-truth depth maps.
}
\vspace{-0.4cm}
\label{vis-nus}
\end{figure*}

In this section, we experimentally validate MMD-LoRA and analyze each of its components for adverse condition depth estimation on the 
nuScenes dataset~\cite{caesar2020nuscenes} and Oxford RobotCar dataset~\cite{maddern20171},
where includes some ablation studies and the comparison against other state-of-the-art depth estimators. 
Then, some qualitative results of MMD-LoRA are shown on the nuScenes dataset and RobotCar dataset.
In the Supplementary, we conduct the influence 
of other hyper-parameters in MMD-LoRA.

\begin{table*}[t]
\scriptsize
\centering
\caption{Evaluation on the RobotCar test set. 
The bold entries denote that our MMD-LoRA surpasses previous SOTA results.
}
\vspace{-0.2cm}
\renewcommand{\arraystretch}{1.3}
\setlength{\tabcolsep}{3.0mm}{
\begin{tabular}{l c c c c c c c c}
\toprule
\multirow{2}{*}{Method} & \multicolumn{4}{c}{day-RobotCar} & \multicolumn{4}{c}{night-RobotCar} \\
\cmidrule(r){2-5} \cmidrule(r){6-9}
& absREL($\downarrow$) & sqREL($\downarrow$) & RMSE($\downarrow$) & $d_1$($\uparrow$) & absREL($\downarrow$)& sqREl($\downarrow$) & RMSE($\downarrow$) & $d_1$($\uparrow$) \\ 
\midrule
Monodepth2~\cite{godard2019digging}		&0.1196 	&0.670 &	3.164 &	86.38 	&0.3029 	&1.724 &	5.038&	45.88 \\
DeFeatNet~\cite{defeat}		&0.2470 	&2.980 	&7.884 	&65.00 	&0.3340 	&4.589 	&8.606	&58.60 \\
ADIDS~\cite{liu2021self}		&0.2390 	&2.089 	&6.743 	&61.40 	&0.2870 	&2.569 	&7.985	&49.00 \\
RNW~\cite{yang2024self}		&0.2970 	&2.608 &7.996 	&43.10 &	0.1850 	&1.710 	&6.549	&73.30 \\
WSGD~\cite{vankadari2023sun}	&0.1760 	&1.603 	&6.036 &	75.00 	&0.1740 	&1.637 	&6.302	&75.40 \\
md4all-DD~\cite{md4all}	&0.1128 &	0.648 	&3.206 &	87.13 	&0.1219 	&0.784 	&3.604	&84.86 \\
DPT with ft.~\cite{tosi2024diffusion} &0.1230 &0.724 &3.333 &86.62& 0.1330 &0.824 &3.712 &83.95\\
DepthAnything with ft.~\cite{tosi2024diffusion}  &0.1190 &0.728 &3.287 &87.17 &0.1290& 0.751 &3.661 &83.68\\
MMD-LoRA(ours) 	&\textbf{0.0796} 	&\textbf{0.324} &\textbf{2.191} 	&\textbf{92.56} &	\textbf{0.0881} 	&\textbf{0.385} 	&\textbf{2.643}&	\textbf{89.33}\\ 
\bottomrule
\end{tabular}}
\label{main_tab_robort}
\vspace{-0.2cm}
\end{table*}
\begin{figure*}[t]
\begin{center}

\includegraphics[width=0.9\textwidth]{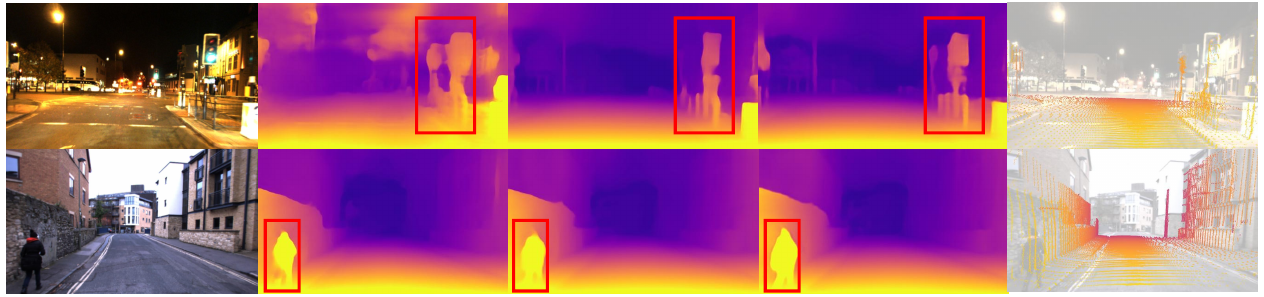}
\end{center}
\vspace{-0.5cm}
\caption{Qualitative results of MMD-LoRA and the previous SOTA depth estimation method on the Robotcar test set. 
The first column denotes the original image. The second, third and fourth column denote the depth estimation results of Monodepth2, md4allDD and ours MMD-LoRA respectively. 
The final column indicates the ground-truth depth maps.
}
\vspace{-0.2cm}
\label{vis-rbot}
\end{figure*}

\begin{table}[!t]
\scriptsize
\centering
\renewcommand{\arraystretch}{1.3}
\caption{
Ablation study of components on the nuScenes validation set. 
}
\vspace{-0.3cm}
\setlength{\tabcolsep}{1.35mm}{
\begin{tabular}{c c c c c c c c}
\toprule
\multirow{2}{*}{PDDA} &\multirow{2}{*}{VTCCL} & \multicolumn{2}{c}{day-clear} & \multicolumn{2}{c}{night} & \multicolumn{2}{c}{day-rain} \\ 
\cmidrule(r){3-4}  \cmidrule(r){5-6} \cmidrule(r){7-8}
& & RMSE($\downarrow$) & d1($\uparrow$)  & RMSE($\downarrow$) & d1($\uparrow$) & RMSE($\downarrow$) & d1($\uparrow$) \\ 
\midrule
\texttimes & \texttimes & 3.214	&95.70 	&8.260 	&74.49	&3.539	&94.52 \\ 
\texttimes & \checkmark	&\textbf{3.118}	&96.09 &7.742	&\textbf{80.46}	&3.565	&94.09 \\ 
\checkmark & \checkmark &3.192	&\textbf{96.46}	&\textbf{7.127}	&79.96	&\textbf{3.417}	&\textbf{95.37}\\ \bottomrule
\end{tabular}
\label{ablation_tab_nusence}}
\vspace{-0.4cm}
\end{table}

\begin{figure*}[!t]
\centering
\includegraphics[width=0.9\textwidth]{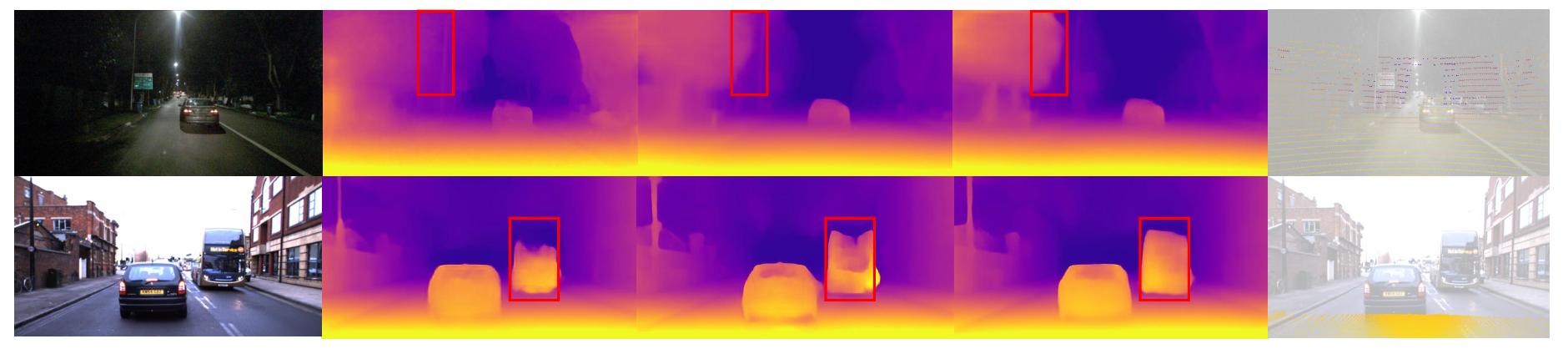}
\vspace{-0.3cm}
\caption{Ablation visualization of our proposed MMD-LoRA with PDDA and VTCCL.
The first column denotes the original image. The second, third and fourth column denote baseline(\ie EVP~\cite{EVP}), MMD-LoRA with PDDA, MMD-LoRA with PDDA and VTCCL. 
The final column indicates the ground-truth depth maps.
}
\label{vis-xiaorong}
\vspace{-0.4cm}
\end{figure*}

\subsection{Datasets and Evaluation Metrics}

We conduct experiments on two challenging and widely used benchmarks under challenging condition depth estimation: nuScenes dataset~\cite{caesar2020nuscenes} and Oxford RobotCar dataset~\cite{maddern20171}. 
For the \textbf{nuScenes dataset}, it is a challenging large-scale dataset with 1000 scenes driving in Boston and Singapore known for its diverse weather conditions and integration of LiDAR data. We adopt the split recommended in md4all~\cite{md4all}, which contains 15129 training images and 6019 validation images. Note that the training images only include day-clear images without adverse weather images and validation images are categorized into day-clear, night and day-rain weather conditions.
For the \textbf{Oxford RobotCar dataset}, it was collected in Oxford, UK, by traversing the same route multiple times in a year, which features a mix of day and night scenes. We adopt the split recommended in md4all~\cite{md4all}, which contains 16563 training images(with day-clear and without night weather) and 1411 validation images(with day-clear and night weather). 
We report on the standard metrics and errors up to 50m for RobotCar and 80m for nuScenes, where these standard metrics are defined in the Supplementary.







\subsection{Implementation Details}

The proposed MMD-LoRA is trained during the pre-trained step by PDDA and VTCCL. 
The baseline depth estimator(\ie EVP) with frozen MMD-LoRA is trained during the training step.
During the pre-trained step, we randomly crop the images into 15 400 $\times$ 400 batch and set base learning rate 0.001, train the MMD-LoRA for 4000 iterations with the batch size of 4 using AdamW optimizer. 
We set the weight type as $W_q$,$W_k$,$W_v$,$W_{proj}$ and the rank r=8 in the MMD-LoRA to adapt to unseen target-domain(\ie adverse weather).
In the VTCCL, we set $\lambda_{0}$: $\lambda_{1}$: $\lambda_{2}$ = 1:0.1:1 for nuScenes dataset and $\lambda_{0}$: $\lambda_{1}$: = 1:0.05 for Oxford RobotCar dataset as the contrastive learning weight coefficients under different weather conditions. 
During the training step, we select the EVP~\cite{EVP} depth estimator as our baseline, the other training settings in the MMD-LoRA method are adopted from the setups in EVP~\cite{EVP}.
In all our experiments, we run the publicly available deep learning framework PyTorch on 2 NVIDIA L20 GPUs.

\subsection{State-of-the-Art Comparison}

We compare the proposed method with the leading monocular depth estimation models under adverse conditions, \ie, \cite{godard2019digging, guizilini20203d, gasperini2021r4dyn, yang2024self, adabins, md4all, yang2024depth, ranftl2020towards,tosi2024diffusion}. Comparison results on the nuScenes dataset are shown in Table~\ref{main_tab_nus}, we can observe that our MMD-LoRA method surpasses all previous approaches, achieving the best results on the majority of the metrics. 
Compared with recent SOTA, \ie \cite{md4all}, our method is better by a large margin. For example, our MMD-LoRA exceeds the previous state-of-the-art method by 8.43\% (from 88.03\% to 96.46\%) in the clear day, 4.63\%(from 75.33\% to 79.96\%) in the night, 12.55\%(from 82.82\% to 95.37\%) in the rainy day in $d_1$. 
The comparison results demonstrate the effectiveness of our proposed MMD-LoRA. Unlike the md4all~\cite{md4all} and Fabio~\etal~\cite{tosi2024diffusion} work, these works rely on generative models to convert the sunny condition into adverse weather and further capture target-domain visual representations. The MMD-LoRA becomes more straightforward and yet more potent in capturing target-domain visual features.
As shown in Fig~\ref{vis-nus}, our method can identify critical elements of the scenes in the night and sunny condition, and adjacent objects or the same object are more distinguishable. For example, our MMD-LoRA correctly captures the complete obstacle, clearly distinguishes two pillars and recovers the 'holes' in the car.

To demonstrate the robustness of our MMD-LoRA, we also evaluate the MMD-LoRA on the RobotCar dataset. The results are shown in Table~\ref{main_tab_robort}, compared with the state-of-the-art methods like~\cite{godard2019digging, defeat, liu2021self,yang2024self,vankadari2023sun,md4all,tosi2024diffusion}. Similar to the above experiments, it outperforms all previous methods. Compared with previous SoTA~\cite{tosi2024diffusion} work, our method successfully improves $d_1$ from 87.17\% to 92.56\% in the day and from 83.68\% to 89.33\% in the night. 
As shown in Fig.~\ref{vis-rbot}, our method clearly delivered the sharp estimation for the standing pillar and human head in the 1st and 2nd rows night and sunny image. 
The comparison on the RobotCar dataset further proves that MMD-LoRA is effective for adverse condition depth estimation.
Both Table~\ref{main_tab_nus} and Table~\ref{main_tab_robort} also show that our MMD-LoRA can obtain the rain-domain and night-domain visual representation in the real-world scene and the visual representations are robust to any unseen target-domain.


\begin{table*}[!t]
\scriptsize
\centering
\caption{The depth estimation performance in learned augmentation~\cite{vidit2023clip} and learned LoRA~\cite{hu2021lora}.}
\vspace{-0.2cm}
\renewcommand{\arraystretch}{1.3}
\begin{tabular}{c c c c c c c c c c c}
\toprule
\multirow{2}{*}{Method} & \multirow{2}{*}{parameters(M)} & \multicolumn{3}{c}{day-clear-nuScenes} & \multicolumn{3}{c}{night-nuScenes} & \multicolumn{3}{c}{day-rain-nuScenes} \\ 
\cmidrule(r){3-5} \cmidrule(r){6-8} \cmidrule(r){9-11}
&  & absREL($\downarrow$) & RMSE($\downarrow$) & d1($\uparrow$) & absREL($\downarrow$) & RMSE($\downarrow$) & d1($\uparrow$) & absREL($\downarrow$) & RMSE($\downarrow$) & d1($\uparrow$) \\ \midrule
baseline~\cite{EVP} &--- &0.070 &3.214 &95.70 &0.184 &8.260 &74.49& 0.081 &3.539& 94.52\\
learned augment~\cite{vidit2023clip} &+0.604	&0.072&3.312&	96.37	&0.169	&7.967	&76.09	&0.079	&3.642&	94.74 \\
learned LoRA(Ours)~\cite{hu2021lora} & \textbf{+0.035} & \textbf{0.069} & \textbf{3.192} & \textbf{96.46} & \textbf{0.155} & \textbf{7.127} & \textbf{79.96} & \textbf{0.074} & \textbf{3.417} & \textbf{95.37} \\ 
\bottomrule
\end{tabular}
\label{tab:different domain generalization methods}
\end{table*}

\subsection{Ablation Studies}

\begin{table*}[t]
\scriptsize
\centering
\renewcommand{\arraystretch}{1.3}
\caption{The depth estimation performance in different rank r of attention weights in MMD-LoRA during the pre-trained stage on the nuScenes validation set and RobotCar test set. Adapting rank r=8 gives the best performance overall.}
\vspace{-0.3cm}
\setlength{\tabcolsep}{0.4mm}{
\begin{tabular}{c c c c c c c c c c c c c c c c c}
\toprule
\multirow{2}{*}{rank r} 
&\multirow{2}{*}{parameter(M)}
&\multicolumn{3}{c}{day-RobotCar} & \multicolumn{3}{c}{night-RobotCar} &\multicolumn{3}{c}{day-clear-nuScenes} & \multicolumn{3}{c}{night-nuScenes
} &\multicolumn{3}{c}{day-rain-nuScenes} 
\\ \cmidrule(r){3-5} \cmidrule(r){6-8} \cmidrule(r){9-11} \cmidrule(r){12-14} \cmidrule(r){15-17}
&  &absREL($\downarrow$) & RMSE($\downarrow$) & d1($\uparrow$) & absREL($\downarrow$) & RMSE($\downarrow$) & d1($\uparrow$) & absREL($\downarrow$) & RMSE($\downarrow$) & d1($\uparrow$) 
& absREL($\downarrow$) & RMSE($\downarrow$) & d1($\uparrow$)
& absREL($\downarrow$) & RMSE($\downarrow$) & d1($\uparrow$)
\\ 
\midrule
16 &0.068 &0.0778   & 2.193  &    92.53&	0.0887  & 2.695    &89.01	&0.0727  & 3.210  & 95.73	&0.1624 & 7.199   &77.29	&0.085 & 3.617   & 93.83 \\ 
8 & \textbf{0.035} & \textbf{0.0796} & \textbf{2.191} & \textbf{92.56} & \textbf{0.0881} & \textbf{2.643} & \textbf{89.33} & \textbf{0.0690} & \textbf{3.192} & \textbf{96.46} &
\textbf{0.1545} & \textbf{7.127} & \textbf{79.96} &
\textbf{0.074} & \textbf{3.417} & \textbf{95.37} \\ 
4 &0.018	&0.0779   &  2.192      &92.54&	0.0886   &2.693  & 89.09	&0.0646 &3.089  &96.49	&0.1898 & 8.289 & 72.26&	0.074 & 3.463  &95.44 \\ 
2 &0.010	&0.0778   &  2.194  &   92.52	&0.0887  &2.699  &89.01	&0.0688 & 3.185 & 96.02	&0.2007 & 8.786  &69.72	&0.072  & 3.411 & 95.35 \\ 
\bottomrule
\end{tabular}
}
\label{tab:comparison_rank}
\vspace{-0.3cm}
\end{table*}

As shown in Table~\ref{ablation_tab_nusence}, although EVP already performs well under adverse conditions, we can observe that the performance of MMD-LoRA with PDDA still outperforms the EVP by 0.39\% and 5.97\% in $d_1$ under day-clear and night. 
The reason for this improvement is that MMD-LoRA with PPDA indeed captures target-domain visual features without the need for additional target-domain images. 
As shown in the 2nd and 3rd columns of Fig.~\ref{vis-xiaorong}, the EVP tends to induce depth artifacts, particularly when estimating depths at object boundaries. 
Compared with EVP, the MMD-LoRA with PPDA achieves a clear border and removes noise in the background. 
Besides, by introducing the VTCCL into MMD-LoRA, we can see that the MMD-LoRA obtains 96.46\% in $d_1$ and 
the errors remain equivalent or even improve by a large margin compared to only using PDDA. 
The reason is that VTCCL achieves robust multimodal alignment, reinforcing consistent representations.
As shown in the 3rd and 4th columns of Fig.~\ref{vis-xiaorong}, VTCCL further achieves the shaper outline and pixel-level accurate depth estimates with clear object boundaries(\ie tree borders, complete truck), and induces depth artifacts seen as the red box.

As shown in Table~\ref{tab:different domain generalization methods}, we explore the depth estimation performance of learned augmentation-based method~\cite{vidit2023clip} and learned model-based method(\ie MMD-LoRA) for adverse condition depth estimation.
While the learned augmentation-based method~\cite{vidit2023clip} and learned model-based method bring 1.60\% and 5.47\% improvement in the night, 0.22\% and 0.85\% improvement in the day-rain for $d_1$, exceeding the baseline by a large margin. 
Compared to learnable augmentation-based methods, the learned model-based method achieves superior depth estimation performance with only a minimal increase in parameter count (0.035M).
 
We conduct ablation experiments to find a suitable rank $r$ for the specific layers of self-attention in the image encoder. 
Table~\ref{tab:comparison_rank} shows the comparison of depth estimation performance utilizing different rank $r$.
The MMD-LoRA achieves the excellent $d_1$ and the comparable error metrics when the rank $r$ is set to 8 with gradually increasing $r$.
Interestingly, the depth estimation performance does not always get better by adding rank $r$. 
This could be explained by the fact that setting the larger threshold value $r>8$ may lead to the predicted unseen target-domain representations overfitting the corresponding text description and deviate from features in the real world. So we set $r$ to 8 for all the experiments in our paper.

\section{Conclusion}
\label{sec:conclusion}

Existing adverse condition depth estimation methods often require additional target images and a depth estimator, yet they fail to achieve adequate alignment between multimodal features. Our proposed approach, MMD-LoRA, addresses these limitations by parameter-efficient fine-tuning techniques and contrastive learning paradigm. Specifically, Prompt Driven Domain Alignment (PDDA) utilizes trainable low-rank adaptation matrices within the image encoder guided by text embeddings to effectively capture visual features from unseen target domains. Subsequently, Visual-Text Consistent Contrastive Learning (VTCCL) separates embeddings of different weather conditions while bringing similar ones closer together, ensuring robust multimodal alignment and enhancing MMD-LoRA’s generalization across diverse adverse scenarios. Comprehensive quantitative and qualitative experiments validate the effectiveness of MMD-LoRA. Future work will explore extending this approach to video-based applications.


\textbf{Limitations.}
\label{Limitations}
The proposed MMD-LoRA method depends on predefined text descriptions for both source and target domains, which are assumed to be known in advance in most applications. Additionally, the experiments assume that the brightness of a sunny day remains consistent, even under adverse weather conditions (e.g., rainfall during rainy weather or visibility during nighttime). This assumption may not hold in scenarios with fine-grained changes in weather conditions.
{
    \small
    \bibliographystyle{IEEEtran}
    \bibliography{main}
}


\end{document}